\documentclass[11pt]{article}
\usepackage{amssymb}
\usepackage{textcomp}
\makeindex
\newtheorem{thm}{Theorem}[section]

\newtheorem{cor}[thm]{Corollary}
\newtheorem{lem}[thm]{Lemma}
\newcommand{\qed}{\hfill $\blacksquare$}
\newcommand{\rchi}{\mbox{\raisebox{0.04in}{$\chi$}}} 
\newcommand{\ttilde}{\raisebox{0.04in}{\texttildelow}} 
\title{The Exact Schema Theorem}
\author{{\bf Alden H. Wright}\\
  Computer Science Department\\
  University of Montana\\
  Missoula, MT 59812\\
  alden.wright@umontana.edu\\
  http://web-dev.cs.umt.edu/\ttilde wright/wright.htm
}
\begin{document}

\maketitle
\begin{center}
\textbf{Comment added on May 18, 2011:}
This paper was written in 1999 with the last revision done 
on January 28, 2000.  A colleague disuaded me from submitting it for
publication, but it does contain useful results.  It has been published on my 
website (formerly \texttt{http://www.cs.umt.edu/u/wright}) since that 1999.
It has been referenced by other publications.

On May 18, 2011, I reprocessed the LaTex source using LaTex, dvips,
and ps2pdf.  (There were some fractions that did not display correctly.)
Except for the addition of this comment, updating my e-mail and website
URL, and the reprocessing,
there have been no changes since the verion of January 28, 2000.
\end{center}

\begin{abstract}
A schema is a naturally defined subset of the space of fixed-length
binary strings.  The Holland Schema Theorem \cite{holland1975} gives a 
lower bound on the expected fraction of a population in a schema after
one generation of a simple genetic algorithm.  
This paper gives formulas for the exact expected fraction of 
a population in a schema after one generation of the simple genetic 
algorithm.  

Holland's schema theorem has three parts, one for selection, one
for crossover, and one for mutation.  The selection part is exact,
whereas the crossover and mutation parts are approximations.
This paper shows how the crossover and mutation parts can be made
exact.  Holland's schema theorem follows naturally as a corollary.

There is a close relationship between schemata and the representation
of the population in the Walsh basis.  This relationship is used
in the derivation of the results, and can also make computation
of the schema averages more efficient.

This paper gives a version of the Vose infinite population model
where crossover and mutation are separated into two functions
rather than a single ``mixing'' function.

\end{abstract}

\section{Introduction}

Holland's schema theorem \cite{holland1975} has been widely used 
for the theoretical analysis of genetic algorithms.  However, it has two 
limitations.  First, it only gives information for a single generation.
Second, it is an approximation, giving only lower bounds on the
schema frequencies.  This paper removes the second limitation.

Michael Vose and coworkers have introduced exact models of the simple
genetic algorithm.  The Vose infinite population model
exactly describes the expected behavior from one generation to the next.
The Markov chain model is an exact model of the finite population
behavior of the simple GA.

Stephens et. al. \cite{stephens:degrees_freedom} describe
these models as ``fine-grained''.  They can be
used to qualitatively describe certain aspects of the behavior of the i
simple GA.  For example, the fixed points of the infinite population model
can be used to describe phenomena such as punctuated equilibria.
(See \cite{vose_liepins:punctuated} and \cite{vose:rhs} for example.)
However, due to the large size of the models, it is generally
impossible to apply these models quantitatively to practical-sized 
problems.

Thus, as is pointed out in \cite{stephens:degrees_freedom}
and \cite{stephens:schemata_building},
a more coarse-grained version of these models is needed.
Models are needed that describe the behavior of a subset of 
of the variables included in the exact models.  For example,
a higher-level organism may have in the order of magnitude of
100,000 genes.  However, population geneticists generally do
not try to model all of these; instead they may use 1-locus and 
2-locus models.  Modeling using schemata is the equivalent technique
for string-representation genetic algorithms; they model the behavior 
of the GA at a subset of the string positions.

In earlier work, Bridges and Goldberg \cite{bridges_goldberg:reproduction}
derived an exact expression for expected number of copies
of a string under one generation of selection and one-point crossover, 
and they claim that their formulas can be extended to find the
expected number of elements in a schema under the same conditions.  
Their formulas are complex and not particularly illuminating.

As mentioned before, Stephens and coworkers 
(\cite{stephens:degrees_freedom} and \cite{stephens:schemata_building})
have results similar to ours for one-point crossover.  
Our results are more general than these results in that they for 
general crossover, and they
include mutation.  

\cite{stephens:degrees_freedom} includes references to other related 
papers.  Of particular note is \cite{DBLP:conf/foga/Altenberg95} which relates
an exact version of the schema theorem to Price's theorem in 
population genetics.

Chapter 19 of \cite{vose:book} (which the author had not
seen when he wrote this paper) also contains a version of
the exact schema theorem as theorem 19.2 for mixing,
where mixing includes crossover and mutation.  Theorem 19.2
assumes that mutation is independent, which is similar
to the assumptions on mutation in this paper.

\section{Notation}

Let $\Omega$ be the space of length $\ell$ binary strings, and
let $n = 2^\ell$.  For $u,v \in \Omega$, let $u \otimes v$ denote
the bitwise-and of $u$ and $v$, and let $u \oplus v$ denote the
bitwise-xor of $u$ and $v$.  Let $\overline{u}$ denote the
ones-complement of $u$, and $\#u$ denote the number of ones in
the binary representation of $u$.

Integers in the interval $[0,n) = [0,2^\ell)$ 
are identified with the elements of $\Omega$ through their binary 
representation.  This correspondence allows $\Omega$ to be regarded 
as the product group
$$
\Omega = Z_2 \times \ldots \times Z_2
$$
where the group operation is $\oplus$.
The elements of $\Omega$ corresponding to the integers $2^i$,
$i = 0,\ldots,\ell-1$ form a natural basis for $\Omega$.

We will also use column vectors of length $\ell$ to represent
elements of $\Omega$.  
Let $\bf 1$ denote the vector of ones (or the integer $2^\ell-1$). 
Thus, $u^Tv = \#(u \otimes v)$, and $\overline{u} = {\bf 1} \oplus u$.

For any $u \in \Omega$, let $\Omega_u$ denote the subgroup
of $\Omega$ generated by $\langle 2^i : u \otimes 2^i = 2^i \rangle$.
In other words, $v \in \Omega_u$ if and only if $v \otimes u = v$.
For example, if $\ell = 6$, then $\Omega_9 = \{ 0, 1, 8, 9 \}
= \{ 000000, 000001, 001000, 001001 \}$.

A schema is a subset of $\Omega$ where some string positions are
specified (fixed) and some are unspecified (variable).  Schemata are
traditionally denoted by pattern strings, where a special symbol is used 
to denote a unspecified bit.  We use the $*$ symbol for this
purpose (Holland used the $\#$ symbol).  Thus, the schema denoted 
by the pattern string $10\!*\!01*$ is the set of strings $\{ 100010,  
100011, 101010, 101011,\}$.

Alternatively, we can define a schema to be the set $\Omega_u \oplus v$, 
where $u, v \in \Omega$, and where $u \otimes v = 0$.  In this notation, 
$u$ is a mask for the variable positions, and $v$ specifies the fixed 
positions.  For example, the schema $\Omega_{001001} \oplus 100010$ 
would be the schema $10\!*\!01*$ described above.

This definition makes it clear that a schema $\Omega_u \oplus v$ with
$v = 0$ is a subgroup of $\Omega$, and a schema $\Omega_u \oplus v$
is a coset of this subgroup.  

Following standard practice, we will define the {\it order} of
a schema as the number of fixed positions.  In other words, the order
of the schema $\Omega_{\overline{u}} \oplus v$ is $\# u$ (since $u$
is a mask for the fixed positions).

A population for a genetic algorithm over length $\ell$ binary strings 
is usually interpreted as a multiset (set with repetitions) of elements of 
$\Omega$.  A population can also be interpreted as a $2^\ell$ dimensional
incidence vector over the index set $\Omega$: if $X$ is a population vector,
then $X_i$ is the number of occurences of $i \in \Omega$ in the
population.  A population vector can be normalized by dividing by
the population size.  For a normalized population vector $x$, 
$\sum_i x_i = 1$.  Let 
$$
\Lambda = \{ x \in R^n : \sum_i x_i = 1
\mbox{ and } x_i \geq 0 \mbox{ for all } i \in \Omega\}.
$$
Thus a normalized population vector is an element
of $\Lambda$.  Geometrically, $\Lambda$ is the $n-1$ dimensional
unit simplex in $R^n$.  Note that elements of $\Lambda$ can be
interpreted as probability distributions over $\Omega$.

If $expr$ is a Boolean expression, then 
$$
[ expr ] = \left\{
\begin{array}{ll} 
1 & \mbox{~~~if } expr \mbox{ is true }\\
0 & \mbox{~~~if } expr \mbox{ is false }
\end{array} \right.
$$

\section{The fraction of a population in a schema }

Let $X$ be a population (not necessarily normalized).
We will be interested in the fraction $X_k^{(u)}$ of the elements of $X$
that are elements of the schema $\Omega_{\overline{u}}\oplus k$:  
$$
X_k^{(u)} = \frac{\sum_{i \in \Omega_{\overline{u}}} X_{i \oplus k}}
  {\sum_{i \in \Omega} X_{i}}
\mbox{~~~~~~~~~for } k \in \Omega_u.
$$
Note that here $u$ is a mask for the fixed positions of the schema.  

If we divide the numerator and denominator of this fraction by
the population size $r$, and if we let $x = X/r$, then we get
$$
x_k^{(u)} = \frac{\sum_{i \in \Omega_{\overline{u}}} x_{i \oplus k}}
  {\sum_{i \in \Omega} x_{i}}
= \sum_{i \in \Omega_{\overline{u}}} x_{i \oplus k}
$$
In other words, for a normalized population $x$, we use
the notation $x_k^{(u)}$ to denote the {\it schema average} for the
schema $\Omega_{\overline{u}} \oplus k$.
Note that $x_0^{(0)} = 1$ since $\sum_{i \in \Omega} x_i = 1$.

Let $x^{(u)}$ denote the vector of schema averages,
where the vector is indexed over $\Omega_u$.  Note that
$\sum_{v \in \Omega_u} x_v^{(u)} = 1$.  

For a fixed $u$, the family of schemata 
$\{ \Omega_{\overline{u}} \oplus v \: : \: v \in \Omega_u \}$
is called a {\it competing family} of schemata.

\section{The Simple Genetic Algorithm}

The material in this section is mostly taken from 
\cite{vose:book}, \cite{vose:modeling}, and \cite{vose_wright:walshI}.

The simple genetic algorithm can be described through a heuristic
function ${\cal G} : \Lambda \rightarrow \Lambda$.  As we will show
later, $\cal G$ contains all of the details of selection, crossover,
and mutation.  The simple genetic algorithm is given by:

\begin{tabbing}
hhh\=hhh\=hhh\=\kill
1 \> Choose a random population of size $r$ from $\Omega$.\\
2 \> Express the population as an incidence vector $X$ indexed over $\Omega$.\\
3 \> Let $y = {\cal G}(X/r)$.  
  (Note that $X/r$ and $y$ are probability distributions over $\Omega$.)\\
4 \> {\bf for} k from 1 to $r$ {\bf do}\\
5 \>\> Select individual $i \in \Omega$ according to the probability 
     distribution $y$.\\
6 \>\> Add $i$ to the next generation population $Z$.\\
7 \> {\bf endfor}\\
8 \> Let $X = Z$.\\
9 \> Go to step 3.\\
\end{tabbing}

It is shown in \cite{vose:book} that if $X$ is a population, then
$y = {\cal G}(X/r)$ is the expected population after one generation
of the simple genetic algorithm.  Thus, the schema theorem is a statement
about the schema averages of the population $y$.

The heuristic function $\cal G$ can be written as the composition of three
hueristic functions $\cal F$, $\cal C$, and $\cal U$ which describe
selection, crossover, and mutation respectively.  In other words,
${\cal G}(x) = {\cal U}({\cal C}({\cal F}(x))) = {\cal U \circ C \circ F}(x)$.  
Later sections describe each of the three heuristic functions in more detail.

\section{Selection}
\label{sec:selection}

The selection heuristic $\cal F$ for proportional selection is given by:
$$
{\cal F}_k(x) = \frac{f_k x_k}{\sum_{j \in \Omega} f_j x_j}
$$
where $f_k$ denotes the fitness of $k \in \Omega$.

Let $F$ denote the diagonal matrix over $\Omega \times \Omega$ whose
diagonal entries are given by $F_{j,j} = f_j$.  Then the selection 
heuristic can be expressed in terms of matrices by
$$
{\cal F}(x) = \frac{Fx}{{\bf 1}^TFx}
$$

If $X$ is a finite population represented as an incidence vector over
$\Omega$, and if $x = X/r$, then $x_k$ is nonzero
only for those $k$ that are in the population $X$ considered as a multiset.
Thus, the computation of ${\cal F}(x)$ is feasible in practice even for long
string lengths.  Further, the computation of the schema averages
after selection can be done directly from the definition.

\begin{thm}
(Exact schema theorem for proportional selection.)
Let $x \in \Lambda$ be a population, and let $s = {\cal F}(x)$.
Then
$$
s_k^{(u)} = 
\frac{\sum_{j \in \Omega_{\overline{u}}} f_{j \oplus k} x_{j \oplus k}}
{\sum_{k \in \Omega} f_k x_k } 
$$
\end{thm}

We give the following algorithm for computing the schema average
vector $s^{(u)}$ from a finite population $X$.  
Let $I(u) = \{ i \: : \: 0 \leq i < \ell \mbox{ and } u_i = 1 \}$,
where $u_i$ denotes bit $i$ of $u$.
Let $P^{(u)}$ be the function which projects $\Omega$ into $\Omega_u$:
for $j \in \Omega$, let $P_i^{(u)}(j) = j_i$ for $i \in I(u)$.

\begin{tabbing}
hhh\=hhh\=hhh\=\kill\\
{\bf for each } $k \in \Omega_u$ {\bf do}\\
\> $s_k^{(u)} \leftarrow 0$\\
{\bf endfor}\\
{\bf for each } $j \in X$ {\bf do} 
   ~~~~~~~~~~~~~~~~~$\triangleright$ see note below\\
\> $k \leftarrow P^{(u)}(j)$\\
\> $s_k^{(u)} \leftarrow s_k^{(u)} + f_j$\\
{\bf endfor}\\
$\overline{f} \leftarrow 0$\\
{\bf for each } $k \in \Omega_u$ {\bf do}\\
\> $\overline{f} \leftarrow \overline{f} + s_k^{(u)}$ \\
{\bf endfor}\\
{\bf for each } $k \in \Omega_u$ {\bf do}\\
\> $s_k^{(u)} \leftarrow s_k^{(u)}/\overline{f}$ \\
{\bf endfor}\\
{\bf return } $s_k^{(u)}$\\
\end{tabbing}

In this algorithm, the population $X$ is interpreted as a multiset.
Thus, it is assumed that 
``{\bf for each } $j \in X$ {\bf do}'' means that the loop following
is done once for each of the possibly multiple occurences of $j$
in $X$.  In an implementation, it would be useful to identify the
elements of $\Omega_u$ with the integers in the interval $[0,2^{\#u})$,
and to interpret $s^{(u)}$ as a vector indexed over these integers.

Clearly, the complexity of this algorithm is $\Theta(2^{\#u} + rK)$,
where $K$ denotes the complexity of one fitness evaluation.

We now give an example which we will continue through the remaining
sections.

Let $\ell = 5$, $u = 10 = 01010_2$, $r = 5$, $X = \{6,7,10,13,21\}
= \{ 00110, 00111, 01010, 01101, 10101 \}$.  The schema sum vector
is $x^{(10)} = < \frac{1}{5}, \frac{2}{5}, \frac{1}{5}, \frac{1}{5} >$.
Let $f_{6} = 5$, $f_{7} = 3$, $f_{10} = 4$,  $f_{13} = 1$,  $f_{21} = 7$.
This gives $\overline{f} = 20$.  The schema sum vector after selection
is $s^{(10)} = \frac{1}{20}< 7, 8, 1, 4 >$.

\section{Holland's Schema Theorem}

We can now state Holland's Schema theorem \cite{holland1975}.

As in \cite{vose_wright:walshI}, for $u \in \Omega$, define
$$
\mbox{hi}(u) = \left\{
\begin{array}{ll}
0 & \mbox{~~~if } u = 0\\
\max\{i : 2^i \otimes u > 0\} & \mbox{~~~otherwise}
\end{array} \right.
$$
$$
\mbox{lo}(u) = \left\{
\begin{array}{ll}
\ell-1 & \mbox{~~~if } u = 0\\
\min\{i : 2^i \otimes u > 0\} & \mbox{~~~otherwise}
\end{array} \right.
$$
Intuitively, the function $\mbox{hi}(u)$ returns the index high-order
bit of $u$, and $\mbox{lo}(u)$ returns the index of the low-order
bit.  Let  ${\cal L}(u) = \mbox{hi}(u)-\mbox{lo}(u)$.  ${\cal L}(u)$
is often called the defining length of $u$.

\begin{thm}
(Holland's approximate schema theorem.)
Let $x \in \Lambda$ be a normalized population, and let $y = {\cal G}(x)$,
where $\cal G$ includes proportional selection, one-point crossover
with crossover rate $c$, and bitwise mutation with mutation rate $p$. 
Then,
$$
y^{(u)}_v \geq
\frac{\sum_{j \in \Omega_{\overline{u}}} f_{j \oplus k} x_{j \oplus k}}
{\sum_{j \in \Omega} f_j x_j } 
\left(1-c\frac{{\cal L}(u)}{\ell-1}\right)\left(1-p\right)^{\#u}
$$
\end{thm}

\section{The Walsh Basis}

\label{sec:walsh}
The Walsh matrix $W$ has dimension $2^\ell$ by $2^\ell$, 
and has elements defined by
$$
W_{i,j} = 2^{-\ell/2} (-1)^{i^Tj}  =  \frac{1}{\sqrt{n}}(-1)^{i^Tj}
$$
Note that $W$ is symmetric and orthogonal ($WW=I$).  The columns of $W$
define a basis for $R^n$ called the Walsh basis.

As an example, for $\ell = 2$,
$$
W =\frac{1}{2} \left[ \begin{array}{rrrr}
1 & 1 & 1 & 1 \\
1 &-1 & 1 &-1 \\
1 & 1 &-1 &-1 \\
1 &-1 &-1 & 1 
\end{array} \right]
$$

If $x$ is a vector over $\Omega$, then $\widehat{x} = Wx$ can be
interpreted as $x$ written in the Walsh basis, and if $M$ is a 
matrix over $\Omega \times \Omega$, then $\widehat{M} = WMW$
can be interpreted as $M$ written in the Walsh basis.

We are also interested in vectors and matrices indexed over $\Omega_u$.
If $\widehat{x}$ is a vector over $\Omega$ written in the Walsh basis,
let $\widehat{x}_k^{(u)} = 2^{\#\overline{u}/2}\widehat{x}_k$.
Theorem \ref{thm:schema_walsh} will show that $\widehat{x}^{(u)}$
is the Walsh transform of $x^{(u)}$.

We can define a Walsh matrix $W^{(u)}$ indexed over 
$\Omega_u \times \Omega_u$.  For $i, j \in \Omega_u$, define
$$
W_{i,j}^{(u)} = 2^{-\#u/2} (-1)^{i^Tj}
$$

The following theorem shows how the schema sum vector is related
to the Walsh coefficients of the population.

\begin{thm}
For any $u \in \Omega$,
$$
x^{(u)} = W^{(u)} \widehat{x}^{(u)}
$$
\label{thm:schema_walsh}
\end{thm}
{\bf Proof.}
\begin{eqnarray}
(W^{(u)} \widehat{x}^{(u)})_k 
&=&
\nonumber
2^{-\#u/2} \sum_{j \in \Omega_u} (-1)^{j^T k} \widehat{x}_j^{(u)}\\
&=&
\nonumber
2^{\#\overline{u}/2-\#u/2} \sum_{j \in \Omega_u} 
(-1)^{j^T k} \widehat{x}_j^\\
&=&
\nonumber
2^{\#\overline{u}/2-\#u/2}\;2^{-\ell/2} \sum_{j\in\Omega_u} (-1)^{j^Tk}
  \sum_{v\in\Omega} (-1)^{j^Tv} x_v\\
\nonumber
&=&
\label{eqn1}
2^{-\#u} \sum_{v\in\Omega} x_v \sum_{j\in\Omega_u} (-1)^{j^T(v \oplus k)}\\
&=&
2^{-\#u} \sum_{w\in\Omega} x_{w \oplus k} \sum_{j\in\Omega_u} (-1)^{j^Tw}\\
&=&
\label{eqn2}
2^{-\#u}  \sum_{w\in\Omega_{\overline{u}}} x_{w \oplus k} 
    \sum_{j\in\Omega_u} (-1)^{j^Tw}\\
&=&
\nonumber
2^{-\#u}  \sum_{w\in\Omega_{\overline{u}}} x_{w \oplus k} 2^{\#u}\\
&=&
\nonumber
x_k^{(u)}
\end{eqnarray}

To see how equation \ref{eqn2} follows from \ref{eqn1},
note that $\sum_{j\in\Omega_u} (-1)^{j^Tv} = 0$ if $v \in \Omega_u$
and $v \neq 0$.
\qed

Theorem \ref{thm:schema_walsh} shows that the schema averages of 
a family of competing
schemata determine the Walsh coefficients of the population in
a coordinate subspace in the Walsh basis.  To be more specific,
consider $u$ as fixed.  Then $x^{(u)}$ denotes the schema averages
of the family of competing schemata $\Omega_{\overline{u}} + k$,
where $k$ varies over $\Omega_u$.  Theorem \ref{thm:schema_walsh}
shows that these schema averages determine $\widehat{x}^{(u)}$,
which is a rescaling of the projection of $\widehat{x}$ into the
coordinate subspace generated by the elements of $\Omega_u$.

To continue the example started in section \ref{sec:selection},
if $s^{(10)} = \frac{1}{20}< 7,8,1,4 >$, then
$\widehat{s}^{(10)} = W^{(10)}s^{(10)} = \frac{1}{20}< 10, -2, 5, 1 >$.

\section{Crossover}

If parent strings $i, j \in \Omega$ are crossed using a crossover
mask $m \in \Omega$, the children are $(i \otimes m) \oplus 
(j \otimes \overline{m})$ and $(i \otimes \overline{m}) \oplus
(j \otimes m)$.  In the simple genetic algorithm, one child is
chosen randomly from the pair of children.

For each binary string $m \in \Omega$,
let $\rchi_m$ be the probability of using $m$ as a crossover mask.

The {\it crossover matrix} is given by
$$
C_{i,j} = \sum_{m \in \Omega} \frac{\rchi_m + \rchi_{\overline{m}}}{2}
  [i \otimes m \oplus j \otimes \overline{m} = 0]
$$
$C_{i,j}$ is the probability of obtaining $0$ as the result of the
crossover of $i$ and $j$.

Let $\sigma_k$ be the permutation matrix with $i,j$th entry given
by $[j \oplus i = k]$.  Then $(\sigma_k x)_i = x_{i \oplus k}$.
Define the {\it crossover heuristic} ${\cal C} : \Lambda \rightarrow
\Lambda$ by 
$$
{\cal C}_k(x) = (\sigma_k x)^T C \sigma_k x
\mbox{~~~~~~for } k \in \Omega
$$
Corollary 3.3 of \cite{vose_wright:walshI} gives that
the Walsh transform of the crossover matrix $\widehat{C}$ is equal
to $C$.

Vose and Wright \cite{vose_wright:walshI} show that
the $k$th component of ${\cal C}(x)$ with respect to the
Walsh basis is
$$
\sqrt{n} \sum_{i \in \Omega_k} \widehat{x}_i \widehat{x}_{i \oplus k} 
\widehat{C}_{i,i\oplus k}
$$
where $n = 2^\ell$.

\begin{thm}  
(The crossover heuristic in the Walsh basis.)
Let $x \in \Lambda$ and let $\widehat{y}$ denote ${\cal C}(x)$ 
expressed in the Walsh basis.
Then
$$
\widehat{y}_k = \sqrt{n} 
\sum_m \frac{\rchi_m + \rchi_{\overline{m}}}{2}
\widehat{x}_{k \otimes m} \widehat{x}_{k \otimes \overline{m}}.
$$
\label{thm:yhat_crossover}
\end{thm}

{\bf Proof.}
\begin{eqnarray*}
\widehat{y}_k 
&=& 
   \sqrt{n}\sum_{i \in \Omega_k} 
   \widehat{x}_i \widehat{x}_{i \oplus k} C_{i,i \oplus k}\\
&=&
   \sqrt{n}\sum_{i \in \Omega_k} 
   \widehat{x}_i \widehat{x}_{i \oplus k} 
   \sum_m \frac{\rchi_m + \rchi_{\overline{m}}}{2}
   [(i \otimes m = 0) \wedge ((i \oplus k) \otimes \overline{m}=0)]\\
&=&
   \sqrt{n}
   \sum_m \frac{\rchi_m + \rchi_{\overline{m}}}{2}
   \sum_{i \in \Omega_k} 
   \widehat{x}_i \widehat{x}_{i \oplus k} 
   [(i \otimes m = 0) \wedge ((i \oplus k) \otimes \overline{m}= 0)]
\end{eqnarray*}
The condition in the square brackets can only be satisfied 
when $i = k \otimes \overline{m}$, and in this case 
$i \oplus k = (k \otimes \overline{m}) \oplus k = k \otimes m$.
Thus,
\begin{eqnarray*}
\widehat{y}_k 
&=&
   \sqrt{n} 
   \sum_m \frac{\rchi_m + \rchi_{\overline{m}}}{2}
   \widehat{x}_{k \otimes m} \widehat{x}_{k \otimes \overline{m}} 
\end{eqnarray*}
\qed

\begin{thm}
(The crossover heuristic for schema in the Walsh basis.)
Let $x \in \Lambda$ and let $\widehat{y}$ denote ${\cal C}(x)$ 
expressed in the Walsh basis.  Then
$$
\widehat{y}_k^{(u)} = 
\sum_m \frac{\rchi_m+\rchi_{\overline{m}}}{2}\;\;
\widehat{x}_{k \otimes m}^{(u \otimes m)}\;
\widehat{x}_{k \otimes \overline{m}}^{(u \otimes \overline{m})}
$$
\label{thm:walsh_cross_u}
\end{thm}

Proof.
\begin{eqnarray*}
\widehat{y}_k^{(u)}
& = &
2^{\#\overline{u}/2} \widehat{y}_k\\
& = &
2^{\#\overline{u}+ \# u/2} 
\sum_m \frac{\rchi_m+\rchi_{\overline{m}}}{2}\;\;
\widehat{x}_{k \otimes m}\;
\widehat{x}_{k \otimes \overline{m}}\\
& = &
2^{\#\overline{u}+ \# u/2} 
\sum_m \frac{\rchi_m+\rchi_{\overline{m}}}{2}\;
\left(2^{-\#(\overline{ u \otimes m } )/2} \;
\widehat{x}_{k \otimes m}^{(u \otimes m)} \right)
\left( 2^{-\#(\overline{u \otimes \overline{m}})/2}\;
\widehat{x}_{k \otimes \overline{m}}^{(u \otimes \overline{m})}\right)\\
\end{eqnarray*}
Consider the exponents:
\begin{eqnarray*}
-\#(\overline{ u \otimes m } )/2-\#(\overline{u \otimes \overline{m}})/2
& = &
-\ell/2 + \#( u \otimes m )/2 
-\ell/2 + \#( u \otimes \overline{m} )/2 \\
& = &
-\ell + \#u/2
\end{eqnarray*}
Thus,
\begin{eqnarray*}
\widehat{y}_k^{(u)}
& = &
2^{\#\overline{u} + \#u/2} 2^{-\ell + \#u/2} 
\sum_m \frac{\rchi_m+\rchi_{\overline{m}}}{2}\;\;
\widehat{x}_{k \otimes m}^{(u \otimes m)}\;
\widehat{x}_{k \otimes \overline{m}}^{(u \otimes \overline{m})}\\
& = &
\sum_m \frac{\rchi_m+\rchi_{\overline{m}}}{2}\;\;
\widehat{x}_{k \otimes m}^{(u \otimes m)}\;
\widehat{x}_{k \otimes \overline{m}}^{(u \otimes \overline{m})}
\end{eqnarray*}
\qed

To continue the numerical example, suppose that 1-point crossover
with crossover rate $1/2$
is applied to the $\ell=5$ population for which 
$\widehat{s}^{(10)} = \frac{1}{20}< 10, -2, 5, 1 >$.
We want to compute $\widehat{y}_k^{(10)}$ for $k = 0,2,8,10$.  For
$k = 0,2,8$, for every crossover mask $m$, either $k \otimes m = 0$
or $k \otimes \overline{m} = 0$, so 
$\widehat{y}_k^{(10)} = \widehat{s}_k^{(k)}\widehat{s}_0^{(10 \oplus k)}
= \widehat{s}_k^{(10)}$.  

For $k = 10$, there are four possible nontrivial
crossover masks, each with probability $1/8$.  For two of these,
$k \otimes m \neq 0$ and $k \otimes \overline{m} \neq 0$.  This gives
$$
\widehat{y}_{10}^{(10)} 
= \frac{3}{4} \widehat{s}_{10}^{(10)} 
+ \frac{1}{4}  \widehat{s}_{2}^{(2)} 
    \widehat{s}_{8}^{(2)} 
= \frac{3}{4} \widehat{s}_{10}^{(10)} 
+ \frac{1}{4} \left(\sqrt{2} \widehat{s}_{2}^{(10)} \right)
   \left( \sqrt{2} \widehat{s}_{8}^{(10)} \right)
= \frac{3}{4} \cdot\frac{1}{20} + \frac{1}{4} \cdot 2 \cdot
     \frac{-2}{20} \cdot \frac{5}{20}
= \frac{3}{80} - \frac{1}{80}
= \frac{1}{40} 
$$
Thus, $\widehat{y}^{(10)} =  \frac{1}{40}< 20, -4, 10, 1 >$.

The following theorem gives a simple formula for the
exact change in the expected schema averages after crossover.
It is a restatment of theorem ?? of \cite{stephens:degrees_freedom}
and theorem ?? of \cite{stephens:schemata_building}.
It can also be easily derived from theorem 19.2 of \cite{vose:book} 
by setting the mutation rate to be zero.

\begin{thm}
(Exact schema theorem for crossover.)
Let $x$ be a population, and let $y = {\cal C}(x)$.  Then
\begin{equation}
y_k^{(u)} = \sum_m \frac{\rchi_m+\rchi_{\overline{m}}}{2}
  x_{k \otimes m}^{(u \otimes m)}
  x_{k \otimes \overline{m}}^{(u \otimes \overline{m})}
\label{eqn:sch_crossover}
\end{equation}
\label{thm:sch_crossover}
\end{thm}

{\bf Proof.}  
\begin{eqnarray*}
y_k^{(u)}
& = & 
2^{-\#u/2}
\sum_{v \in \Omega_u} (-1)^{k^Tv} \widehat{y}_v^{(u)}\\
& = & 
2^{-\#u/2} 
\sum_{v \in \Omega_u} (-1)^{k^Tv} 
\sum_m \frac{\rchi_m + \rchi_{\overline{m}}}{2} \;\;
\widehat{x}_{v \otimes m}^{(u \otimes m)}\;
\widehat{x}_{v \otimes \overline{m}}^{(u \otimes \overline{m})}\\
& = & 
2^{-\#u/2}
\sum_{i \in \Omega_{u \otimes m}}
\sum_{j \in \Omega_{u \otimes \overline{m}}} 
\sum_m \frac{\rchi_m + \rchi_{\overline{m}}}{2} \;\;
(-1)^{k^T (i \oplus j)}\;
\widehat{x}_{(i \oplus j) \otimes m}^{(u \otimes m)}\;\; 
\widehat{x}_{(i \oplus j) \otimes \overline{m}}^{(u \otimes \overline{m})}\\
& = & 
2^{-\#u/2}
\sum_m \frac{\rchi_m + \rchi_{\overline{m}}}{2} 
\sum_{i \in \Omega_{u \otimes m}} (-1)^{k^T i} \widehat{x}_i^{(u \otimes m)} 
\sum_{j \in \Omega_{u \otimes \overline{m}}} (-1)^{k^T j}
\widehat{x}_j^{(u \otimes \overline{m})}\\
& = & 
2^{-\#u/2}
\sum_m \frac{\rchi_m + \rchi_{\overline{m}}}{2} 
2^{\#(u \otimes m)/2} x_{k \otimes m}^{(u \otimes m)}
2^{\#(u \otimes \overline{m})/2} 
x_{k \otimes \overline{m}}^{(u \otimes \overline{m})}\\
& = & 
\sum_m \frac{\rchi_m + \rchi_{\overline{m}}}{2} 
x_{k \otimes m}^{(u \otimes m)}
x_{k \otimes \overline{m}}^{(u \otimes \overline{m})}\\
\end{eqnarray*}
\qed

Theorems \ref{thm:yhat_crossover}, \ref{thm:walsh_cross_u},
and \ref{thm:sch_crossover} show that the effect of
crossover using mask $m$ is to move the population towards
linkage equilibrium relative to $m$.  Following the
population biologists (see \cite{crow_kimura:pop_gen_theory}
for example), we define
the population $x$ to be in {\it linkage equilibrium}
relative to mask $m$ if $\widehat{x}_k = \widehat{x}_{k \otimes m} 
\widehat{x}_{k \otimes \overline{m}}$, or equivalently if
$x_k^{(u)} = x_{k \otimes m}^{(u \otimes m)}
x_{k \otimes \overline{m}}^{(u \otimes \overline{m})}$
for all $k \in \Omega$.  If a population is in linkage equilibrium 
with respect to all masks of a family of crossover masks that 
separates any pair of bit positions, then the population will be 
completely determined by the order 1 schemata averages (or equivalently 
the Walsh coefficients $\widehat{x}_k$ with $\# k = 1$).
This is formalized in theorem 3.0 of \cite{vose_wright:walshII}
(Geiringer's theorem).

Continuing the numerical example, suppose that one-point crossover 
with crossover rate $1/2$ is applied to the $\ell=5$ population
whose schema averages for $u=10$ are given by 
$s^{(10)} = \frac{1}{20}<7, 8, 1, 4>$.  To apply theorem 
\ref{thm:sch_crossover}, we need $s^{(2)}$ and $s^{(8)}$.
These are easily obtained from $s^{(10)}$:
$s_0^{(2)} = s_0^{(10)} + s_8^{(10)} = \frac{2}{5}$,
$s_2^{(2)} = s_2^{(10)} + s_{10}^{(10)} = \frac{3}{5}$,
$s_0^{(8)} = s_0^{(10)} + s_2^{(10)} = \frac{3}{4}$,
$s_8^{(8)} = s_8^{(10)} + s_{10}^{(10)} = \frac{1}{4}$.

Let $y = {\cal C}(s)$.  As before, the probability of a crossover
mask for which $u \otimes m \neq 0$ and $u \otimes \overline{m} \neq 0$
is $1/4$.  Thus,
\begin{eqnarray*}
y_{0}^{(10)} & = & \frac{3}{4} s_{0}^{(10)} 
     + \frac{1}{4} s_{0}^{(2)} s_{0}^{(8)}
     = \frac{3}{4} \cdot \frac{7}{20} 
       + \frac{1}{4} \cdot\frac{2}{5} \cdot \frac{3}{4}
     = \frac{27}{80}\\
y_{2}^{(10)} & = & \frac{3}{4} s_{2}^{(10)} 
     + \frac{1}{4} s_{2}^{(2)} s_{0}^{(8)}
     = \frac{3}{4} \cdot \frac{8}{20} 
       + \frac{1}{4} \cdot\frac{3}{5} \cdot \frac{3}{4}
     = \frac{33}{80}\\
y_{8}^{(10)} & = & \frac{3}{4} s_{8}^{(10)} 
     + \frac{1}{4} s_{0}^{(2)} s_{8}^{(8)}
     = \frac{3}{4} \cdot \frac{1}{20} 
       + \frac{1}{4} \cdot\frac{2}{5} \cdot \frac{1}{4}
     = \frac{5}{80} = \frac{1}{16}\\
y_{10}^{(10)} & = & \frac{3}{4} s_{10}^{(10)} 
     + \frac{1}{4} s_{2}^{(2)} s_{8}^{(8)}
     = \frac{3}{4} \cdot \frac{4}{20} 
       + \frac{1}{4} \cdot\frac{3}{5} \cdot \frac{1}{4}
     = \frac{15}{80} = \frac{3}{16}
\end{eqnarray*}
One can check that $y^{(10)}$ is the Walsh transform of $\widehat{y}^{(10)}$
computed earlier.

\begin{cor}
(Approximate schema theorem for crossover.)
Let $x$ be a population, and let $y = {\cal C}(x)$.  Then
$$
y_k^{(u)}  \geq
  x_{k}^{(u)}
  \sum_m \frac{\rchi_m+\rchi_{\overline{m}}}{2}
  [(u \otimes m = u) \vee (u \otimes \overline{m} = u)]
$$
\end{cor}
Note that the summation over $m$ includes just those crossover
masks that do not ``split'' the mask $u$.

Proof.  
For $u$ such that $u \otimes m = u$, we have:
\begin{eqnarray*}
u \otimes \overline{m} &=& 0 \mbox{~~~~~~~~~~~~~ and }\\
x^{(u \otimes \overline{m})}_{k \otimes \overline{m}} &= & 
x^{(0)}_0 = 1 \mbox{~~~~~~ and }\\
x^{(u \otimes m)}_{k\otimes m} &=& x^{(u)}_k
\end{eqnarray*}
Similarly, for $u$ such that $u \otimes \overline{m} = u$, we have:
\begin{eqnarray*}
u \otimes m &=& 0\mbox{~~~~~~~~~~~~~ and }\\
x^{(u \otimes m)}_{k \otimes m}&=& x^{(0)}_0 = 1 \mbox{~~~~~~ and }\\
x^{(u \otimes \overline{m})}_{k\otimes \overline{m}} &=& x^{(u)}_k
\end{eqnarray*}
Those terms in the summation of equation (\ref{eqn:sch_crossover})
for which $(u \oplus m = u) \vee (u \oplus \overline{m} = u)$
is not true are nonnegative.  Thus, if we drop those terms from the
summation, we get the equation of the corollary.
\qed

\begin{cor}
(Holland's approximate schema theorem for 1-point crossover.)
Let $x$ be a population, and let $y = {\cal C}(x)$, where
${\cal C}$ is defined through 1-point crossover with a
crossover rate of $c$. Then
$$
y_k^{(u)}  \geq
  x_{k}^{(u)}
  \left(1-c\frac{{\cal L}(u)}{\ell-1}\right)
$$
\end{cor}

Proof.  One-point crossover can be defined using $\ell$ crossover 
masks with a nonzero probability.  The crossover mask $0$ has probability
$1-c$, and the masks of the form $2^i-1$, $i = 1,\ldots,\ell-1$
have probability $c/(\ell-1)$.  The number of crossover masks such
that $(u \otimes m = u) \vee (u \otimes \overline{m} = u)$ is not
true is ${\cal L}(u)$.  Thus, the probability that
$(u \otimes m = u) \vee (u \otimes \overline{m} = u)$ is true is 
$$
  \left(1-c\frac{{\cal L}(u)}{\ell-1}\right)
$$
\qed

It is not hard to give similar approximate schema theorems
for other forms of crossover, such as two-point crossover
and uniform corssover.

\section{Mutation}
In the Vose model, mutation is defined by means of mutation
masks.  If $j \in \Omega$, then the result of mutating $j$
using a mutation mask $m \in \Omega$ is $j \oplus m$.
The mutation heuristic is defined by giving a probability
distribution $\mu \in \Lambda$ over mutation masks.  In other
words, $\mu_m$ is the probability that $m \in \Omega$ is
used.  Given a population $x \in \Lambda$, the mutation
heuristic ${\cal U}:\Lambda \rightarrow \Lambda$ is defined by 
$$
{\cal U}_k(x) = \sum_{j \in \Omega} \mu_{j \oplus k} x_j
$$
The mutation heuristic is a linear operator: it can be defined
as multiplication by the matrix $U$, where 
$U_{j,k} = \mu_{j \oplus k}$.  In other words, 
${\cal U}(x) = Ux$.

In the Walsh basis, the mutation heuristic is represented by
a diagonal matrix.
\begin{lem}
The $k$th component of the mutation heuristic in the Walsh basis
is given by
$$
\sqrt{n} \: \widehat{\mu}_k \: \widehat{x}_k
$$
where $n = 2^{\ell}$.
\label{lem:mut_heur}
\end{lem}

Proof.
It is sufficient to show that the Walsh transform $\widehat{U}$
of $U$ is diagonal since 
$$
\widehat{{\cal U}(x)} = WUx = (WUW)(Wx) = \widehat{U}\widehat{x}
$$
The following shows that $\widehat{U}$ is diagonal.
\begin{eqnarray*}
\widehat{U}_{j,k} 
&=&
\frac{1}{n} \sum_{v,w} (-1)^{j^Tv} (-1)^{k^Tw} U_{v,w}\\
&=&
\frac{1}{n} \sum_v \sum_w (-1)^{j^Tv+k^Tw} \mu_{v \oplus w}\\
\end{eqnarray*}
We now do a change of variable.  Let $u = v \oplus w$, which
implies that $w = v \oplus u$. 
\begin{eqnarray*}
\widehat{U}_{j,k} 
&=&
\frac{1}{n} \sum_v \sum_u (-1)^{j^Tv+k^T(v \oplus u)} \mu_u\\
&=&
\frac{1}{n} \sum_v \sum_u (-1)^{(j \oplus k)^Tv+k^Tu} \mu_u\\
&=&
\frac{1}{n} \sum_u (-1)^{k^Tu} \mu_u \sum_v (-1)^{(j \oplus k)^Tv} \\
&=&
 \sum_u (-1)^{k^Tu} \mu_u \\
&=&
\sqrt{n} \widehat{\mu}_k [j = k]
\end{eqnarray*}
\qed

Define $\mu_k^{(u)} = \sum_{j \in \Omega_{\overline{u}}} \mu_{k \oplus j}$.
Theorem \ref{thm:schema_walsh} shows that 
$\mu^{(u)} = W^{(u)} \widehat{\mu}^{(u)}$, where 
$\widehat{\mu}_k = 2^{\# \overline{u}/2}\widehat{\mu}_k$ for
all $k \in \Omega_u$.

Define the $2^{\#u}\times 2^{\#u}$ matrix $U^{(u)}$ by
$U^{(u)}_{j,k} = \mu^{(u)}_{j \oplus k}$.  Note that $U = U^{({\bf 1})}$.
The proof of lemma \ref{lem:mut_heur} shows that the Walsh transform
$\widehat{U^{(u)}}$ of $U^{(u)}$ is diagonal and 
$\widehat{U^{(u)}}_{k,k} = 2^{\# u/2} \widehat{\mu}_k$.  
Thus, it is consisitent to write $\widehat{U}^{(u)}$ for
$\widehat{U^{(u)}}$.

We now assume that each string position $i$, $i = 0,1,\ldots,\ell-1$, 
is mutated independently of other positions:  with a probability
of $p_i$, the bit at position $i$ is flipped.  If all of the
$p_i$ are equal to a common value $p$, then $p$ is called the
{\it mutation rate}.

Under this assumption, the probability distribution for
mutation masks is given by
\begin{equation}
\mu_m = \prod_{i=0}^{\ell-1} p_i^{m_i} (1-p_i)^{1-m_i}
\label{eqn:mut_ind}
\end{equation}
where $m_i$ denotes bit $i$ of $m$, and where $0^0$ is interpreted 
to be $1$.
For example, the distribtuion for $\ell = 2$ is the vector
$$
< \begin{array}{cccc} 
(1-p_0)(1-p_1) & p_0(1-p_1) & (1-p_0)p_1 & p_0 p_1
\end{array} >^T
$$

We now want to show that there is an equation similar to (\ref{eqn:mut_ind})
for $\mu_m^{(u)}$.  The next lemma is a step in that direction.
For $u \in \Omega$, define
$I(u) = \{i \; : \; 0 \leq i < \ell \mbox{ and } u_i =1 \}$.

\begin{lem}
For $v \in \Omega$,
\begin{equation}
\sum_{k \in \Omega_u} \prod_{i \in I(u)} p_i^{k_i}(1-p_i)^{1-k_i} = 1
\label{eqn:sum_prod}
\end{equation}
\end{lem}
Proof.  The proof is by induction on $\#u$.  

If $\#u = 1$, then $u = 2^j$ for some $j$, and $I(u) = j$.  Also,
$\Omega_u = \{0,u\}$.  Thus, the left side of equation (\ref{eqn:sum_prod})
is $p_j^0(1-p_j)^1 + p_j^1(1-p_j)^0 = (1-p_j) + p_j = 1$.

If $\#u > 1$, let $u = v \oplus w$ with $v \otimes w = 0$, $\#v > 0$,
$\#w > 0$.  Then
\begin{eqnarray*}
\sum_{k \in \Omega_u} \prod_{i \in I(u)} p_i^{k_i}(1-p_i)^{1-k_i}
& = &
\sum_{k \in \Omega_v} \sum_{r \in \Omega_w}
\left( \prod_{i \in I(v)} p_i^{k_i}(1-p_i)^{1-k_i} \right)
\left( \prod_{j \in I(w)} p_j^{r_j}(1-p_j)^{1-r_j} \right)\\
& = &
\left( \sum_{k \in \Omega_v} 
\prod_{i \in I(v)} p_i^{k_i}(1-p_i)^{1-k_i} \right)
\left( \sum_{r \in \Omega_w} 
\prod_{j \in I(w)} p_i^{r_j}(1-p_j)^{1-r_j} \right)\\
& = & 1
\end{eqnarray*}
\qed

\begin{lem}
For $u \in \Omega$ and $m \in \Omega_u$,
\begin{equation}
\mu_m^{(u)} = \prod_{i \in I(u)} p_i^{m_i} (1-p_i)^{1-m_i}
\label{eqn:mut_dist}
\end{equation}
\end{lem}

Proof.
\begin{eqnarray*}
\mu_m^{(u)}
& = &
\sum_{v \in \Omega_{\overline{u}}} \mu_{m \oplus v}\\
& = &
\sum_{v \in \Omega_{\overline{u}}}
\left( 
\prod_{i \in I(u)} p_i^{(m \oplus v)_i}(1-p_i)^{1-(m \oplus v)_i}
\right)
\left( 
\prod_{j \in I(\overline{u})} p_j^{(m \oplus v)_j}(1-p_j)^{1-(m \oplus v)_j}
\right)\\
& = &
\left( \prod_{i \in I(u)} p_i^{m_i}(1-p_i)^{1-m_i} \right)
\left(
\sum_{v \in \Omega_{\overline{u}}}\;\;
\prod_{j \in I(\overline{u})} 
p_j^{(m \oplus v)_j}
(1-p_j)^{1-(m \oplus v)_j}
\right)
\end{eqnarray*}
Do a change of variable:  let $w = v \oplus (m \otimes \overline{u})$.
Then 
\begin{eqnarray*}
\sum_{v \in \Omega_{\overline{u}}}\;\;
\prod_{j \in I(\overline{u})} p_j^{(m \oplus v)_j}(1-p_j)^{1-(m \oplus v)_j}
& = &
\sum_{w \in \Omega_{\overline{u}}}\;\;
\prod_{j \in I(\overline{u})} p_j^{w_j}(1-p_j)^{1-w_j}\\
& = & 1
\end{eqnarray*}
\qed

The next step is to compute the Walsh transform of the mutation
probability distribution under this assumption.  It is helpful
to do a change of coordinates.  For each $i = 0,1,\ldots,\ell-1$,
let $q_i = 1-2p_i$.  
Under this change of coordinates, equation (\ref{eqn:mut_dist}) is
equivalent to
$$
\mu_m^{(u)} = 2^{-\#u}\prod_{i \in I(u)}\left( 1 + (1-2m_i)q_i\right)
$$

\begin{lem} 
For $m \in \Omega_u$,
$$
\widehat{\mu}_m^{(u)} = 2^{-\#u/2} \prod_{i \in I(m)} q_i
$$
\label{lem:mut_hat_u}
\end{lem}

Proof.
The proof is by induction on $\#u$.  For the base case,
assume that $\#u = 1$.  Then $u = 2^i$ for some $i$, and
\begin{eqnarray*}
\widehat{\mu}^{(u)} &=& W^{(u)} \mu^{(u)}\\
&=&
\frac{1}{\sqrt{2}} \left[ \begin{array}{cc}
1 & 1 \\ 1 & -1 \end{array} \right]
\frac{1}{2} \left[ \begin{array}{c} 1 + q_i \\ 1 - q_i \end{array} \right]\\
&=&
\frac{1}{\sqrt{2}} 
 \left[ \begin{array}{c} 1 \\ q_i \end{array} \right]\\
\end{eqnarray*}
For $\#u > 1$, we have
\begin{eqnarray*}
\widehat{\mu}_m^{(u)}
&=&
2^{-\#u/2}
   \sum_{ j \in \Omega_u} (-1)^{m^Tj} 
   \;\; 2^{-\#u}\prod_{ i \in I(u) }
   \left( 1 + (1-2j_i)q_i\right)\\
\end{eqnarray*}
Let $u = v \oplus w$ where $v \otimes w = 0$, $v \neq 0$,
and $w \neq 0$.  
\begin{eqnarray*}
\widehat{\mu}_m^{(u)}
&=&
2^{-\frac{3}{2}\#(v \oplus w)}
\sum_{ j \in \Omega_{v \oplus w}} (-1)^{m^Tj} \prod_{ i \in I(v \oplus w) }
\left( 1 + (1-2j_i)q_i\right)\\
&=&
\left(
2^{-\frac{3}{2}\#v}
\sum_{ j \in \Omega_{v}} (-1)^{(m \otimes v)^Tj} \prod_{ i \in I(v) }
\left( 1 + (1-2j_i)q_i\right)
\right)\\
&&
\;\;\;\;\;\;\;\;\;\;\;\left(
2^{-\frac{3}{2}\#w}
\sum_{ j \in \Omega_{w}} (-1)^{(m \otimes w)^Tj} \prod_{ i \in I(w) }
\left( 1 + (1-2j_i)q_i \right)
\right)\\
&=&
\left( \widehat{\mu}_{m \otimes v}^{(v)} \right)
\left( \widehat{\mu}_{m \otimes w}^{(w)} \right)\\
&=&
\left(2^{-\#v/2} \prod_{i \in I(m \otimes v)} q_i\right)
\left(2^{-\#w/2} \prod_{i \in I(m \otimes w)} q_i\right)\\
&=&
2^{-\#u/2} \prod_{i \in I(m)} q_i
\end{eqnarray*}
\qed

\begin{lem}
$$
\widehat{\mu}_m  =\frac{1}{\sqrt{n}} \prod_{i \in I(m)} q_i
$$
\label{lem:mut_hat}
\end{lem}
Proof.  
$$
\widehat{\mu}_m = \widehat{\mu}_m^{({\bf 1})}
 = 2^{\#{\bf 1}/2} \prod_{i \in I(m)} q_i
 = \frac{1}{\sqrt{n}} \prod_{i \in I(m)} q_i
$$
\qed

\begin{thm}
(The mutation hueristic in the Walsh Basis.)
Let $x \in \Lambda$ be a population, and let $y = {\cal U}(x)$.  
If $k \in \Omega_u$, then
$$
\widehat{y}_k^{(u)} = \widehat{x}_k^{(u)} \prod_{i \in I(k)} q_i 
$$
\label{th:walsh_mutation}
\end{thm}

Proof.
\begin{eqnarray*}
\widehat{y}_k^{(u)}
& = &
2^{\#\overline{u}/2} \widehat{y}_k\\
& = &
2^{\#\overline{u}/2} 2^{\ell/2} \widehat{\mu}_k \widehat{x}_k 
\mbox{~~~~~~~~~by lemma \ref{lem:mut_heur}}\\
& = &
2^{\#\overline{u}/2} 2^{\ell/2} 
\left(2^{-\#\overline{u}/2} \mu_k^{(u)} \right)
\left(2^{-\#\overline{u}/2} x_k^{(u)} \right)\\
& = &
2^{\#u/2}\left( 2^{-\#u/2}\prod_{i \in I(k)} q_i \right)
\widehat{x}_k^{(u)}\\
& = &
\widehat{x}_k^{(u)}\;\prod_{i \in I(k)} q_i
\end{eqnarray*}
\qed

Theorem \ref{th:walsh_mutation} shows how mutation affects a
population.  If $k \neq 0$, and if for every $i$, $0 < p_i \leq 1/2$,
then $\prod_{i \in I(k)} q_i < 1$.  Thus, $|\widehat{y}_k^{(u)}|
< |\widehat{x}_k^{(u)}|$.  Mutation is decreasing the magnitude
of the schema Walsh coefficients (except for the index 0 coefficient 
which is constant at $2^{-\# u/2}$).  If all of these Walsh
coefficients were zero, then Theorem \ref{thm:schema_walsh} shows
that all of the corresponding schema averages would be equal.
In other words, mutation drives the population towards uniformity.

To continue the numerical example, we take apply mutation with
a mutation rate of $1/8$ to the population $y$ of the previous section.
We start with $\widehat{y}^{(10)} =  \frac{1}{40}< 20, -4, 10, 1 >$.
Let $z = {\cal U}(y)$.
For all $i$, $q = q_i = 1 - 2p_i = 1 - 1/4 = 3/4$.  Thus,
\begin{eqnarray*}
\widehat{z}_0^{(10)} &=& \widehat{y}_0^{(10)} = \frac{1}{2}\\
\widehat{z}_2^{(10)} &=& \widehat{y}_2^{(10)}\cdot q 
   = \frac{-1}{10} \cdot \frac{3}{4} = - \frac{3}{40}\\
\widehat{z}_8^{(10)} &=& \widehat{y}_8^{(10)}\cdot q 
   = \frac{1}{4} \cdot \frac{3}{4} =  \frac{3}{16}\\
\widehat{z}_{10}^{(10)} &=& \widehat{y}_{10}^{(10)}\cdot q^2 
   = \frac{1}{40} \cdot \frac{9}{16} =  \frac{9}{640}
\end{eqnarray*}

\begin{lem}
For $u \in \Omega$,
$$
\widehat{y}^{(u)} = \widehat{U}^{(u)} \widehat{x}^{(u)}
$$
\end{lem}
Proof.  This is just a rewriting of the equation of theorem
\ref{th:walsh_mutation} into matrix form. \qed

The following theorem can be easily derived from theorem
19.2 of \cite{vose:book} by setting the crossover rate to be zero.

\begin{thm}
(The exact schema theorem for mutation.)
Let $x \in \Lambda$ be a population, and let $y = {\cal U}(x)$
where $\cal U$ corresponds to mutating bit $i$ with probability $p_i$
for $i = 0,1,\dots \ell-1$.  Then
\begin{equation}
y^{(u)} = U^{(u)} x^{(u)}
\label{eqn:exact_schema_mutation}
\end{equation}
\label{th:exact_schema_mutation}
\end{thm}
Proof.
\begin{eqnarray*}
y^{(u)}
& = &
W^{(u)} \widehat{y}^{(u)}\\
& = &
W^{(u)} \widehat{U}^{(u)} \widehat{x}^{(u)}\\
& = &
W^{(u)} (W^{(u)}  U^{(u)} W^{(u)}) W^{(u)} x^{(u)}\\
& = &
U^{(u)} x^{(u)}
\end{eqnarray*}
\qed

We continue the numerical example.  We start with the schema
averages computed in the crossover section:
$y^{(10)} =\frac{1}{80} < 27, 33, 5, 15 >$ and let $z = {\cal U}(y)$
where $\cal U$ corresponds to mutation with a mutation rate of $1/8$.
Recall that $\mu^{(u)}$ is given by equation (\ref{eqn:mut_dist}), so 
$$
\mu^{(10)} = < (1-p)^2, p(1-p), (1-p)p, p^2 > 
= \frac{1}{64}< 49, 7, 7, 1 >
$$  
The entries of $U^{(u)}$ are given
by $U_{j,k}^{(u)}= \mu_{j \oplus k}^{(u)}$, so
\begin{eqnarray*}
z^{(10)} = 
U^{(10)} y^{(10)} = \frac{1}{64} \left[ \begin{array}{rrrr}
49 & 7 & 7 & 1\\
7 & 49 & 1 & 7\\
7 & 1 & 49 & 7\\
1 & 7 & 7 & 49\\
\end{array} \right]
\cdot \frac{1}{80}
\left[ \begin{array}{r} 27 \\ 33 \\ 5\\ 15 \end{array} \right]
= \frac{1}{1280} 
\left[ \begin{array}{r} 401 \\ 479 \\ 143\\ 257 \end{array} \right]
\end{eqnarray*}

\begin{cor}
(The approximate schema theorem for mutation.)
Let $x \in \Lambda$ be a normalized population, and let
$y = {\cal U}(x)$.  Assume that ${\cal U}$ corresponds to mutation where 
each bit is mutated (flipped) with probability $p$. Then 
$$
y_k^{(u)} \geq (1-p)^{\# u} x_k^{(u)}
$$
\end{cor}

Proof.
The diagonal entries of $U^{(u)}$ are all equal to
$\widehat{\mu}_0^{(u)} = \prod_{i \in I(u)} (1-p_i)$.
Under the assumption of this corollary, $p_i = p$ for all $i$,
so the diagonal entries of $U^{(u)}$ are all equal to 
$(1-p)^{\# u}$.  The off-diagonal entries of $U^{(u)}$ are all 
nonnegative.  If we drop the off-diagonal entries in 
the computation of equation (\ref{eqn:exact_schema_mutation}),
we get the result of this corollary.  \qed

\section{Computational Complexity}

In this section we give the computational complexity of computing
the schema averages for a family of competing schema averages after
one generation of the simple GA.

It is more efficient to compute the schema averages after selection
using the normal basis using the algorithm given in section
\ref{sec:selection}, convert to the Walsh basis using the 
Fast Walsh transform (see Appendix A), 
compute the effects of crossover and mutation
in the Walsh basis, and convert back to normal coordinates using
the fast Walsh transform.  To convert from $x^{(u)}$ to 
$\widehat{x}^{(u)}$ by the fast Walsh transform has complexity
$\Theta( \#u \cdot 2^{\#u} )$ (\cite{vose:book}).  The complexity of the 
computation of theorem \ref{thm:walsh_cross_u} is 
$\Theta( \#u \cdot 2^{\#u})$ for one or two point crossover 
(since the summation over $m$ is $\Theta(\#u)$).  
The complexity of the computation
of theorem \ref{th:walsh_mutation} is also $\Theta( \#u \cdot 2^{\#u})$.
Thus, the overall computational complexity (assuming an initial
finite population and one or two point crossover) 
is $\Theta( \# u \cdot 2^{\#u} + rK)$ where $K$
was defined as the cost of doing one function evaluation.
Note that the only dependence on the string length is through $K$.
Thus, it is possible to compute schema averages exactly for
very long string lengths.

\section{Conclusion}

We have given a version of the Vose infinite population model
where the crossover heuristic function and the mutation heuristic
function are separate functions, rather than combined into a
single mixing heuristic function.

We have shown how the expected behavior of a simple genetic
algorithm relative to a family of competing schemata can be computed
exactly over one generation.  

As was mentioned in section \ref{sec:walsh},
these schema averages over a family of competing schemata 
correspond to a coordinate subspace of 
$\Lambda$ as expressed in the Walsh basis.  In \cite{vose_wright:walshI}, 
it was shown that the mixing (crossover and mutation) heuristic is 
invariant over coordinate subspaces in the Walsh basis.  We have
explicitly shown how the Vose infinite population model (the heuristic
function $\cal G$) can be computed on these subspaces.  In fact,
the model works in essentially the same way on schema averages
as it does on individual strings.

The formulas are simply stated and easy to understand.  They are 
computationally feasible to apply even for very long string lengths 
if the order of the family of competing schemata is small.
(The formulas are exponential in the order of the schemata.)

A result like the exact schema theorem is most useful if it can
be applied over multiple generations.  The results of this paper
show that the obstacle to doing this is selection, rather than
crossover and mutation.  The result of the exact schema theorem
is the exact schema averages of the family of competing schemata (or
the corresponding Walsh coefficients) after one generation.  
These correspond to an
``infinite population'' which has nonzero components over all elements
of $\Omega$.  If the string length is long and no assumptions are
made about the fitness function, then the effect of selection
on the schema averages for the next generation will be computationally 
infeasible to compute.
Thus, in order to apply the exact schema theorem over multiple generations
for practically realistic string lengths, one will have to make 
assumptions about the fitness function.  A subsequent paper will 
explore this problem.

{\bf\large Acknowledgements:}  The author would like to thank
Yong Zhao, who proofread a version of this paper.

\bibliographystyle{alpha}
\bibliography{genetic}

\clearpage
{\bf \Large Appendix: Table of Notation}

\begin{tabular}{ll}
$[ e ]$ & = 1 if $e$ is true, 0 if $e$ is false\\[0.07in]
$\ell$ & The string length \\[0.07in]
$c$ & The arity of the alphabet used in the string representation\\[0.07in]
$\Omega$ & The set of binary strings of length $\ell$\\[0.07in]
$n$ & $= c^\ell$, the number of elements of $\Omega$\\[0.07in]
$r$ & The population size \\[0.07in]
$u \oplus v$ & The strings $j$ and $k$ are bitwise added mod 2, 
   (or bitwise XORed)\\[0.07in]
$u \otimes v$ & The strings $j$ and $k$ are bitwise multiplied mod 2, 
   (or bitwise ANDed)\\[0.07in]
$\overline{u}$ & The ones complement of the string $j$\\[0.07in]
$\#u$ & The number of ones in the binary string $u$\\[0.07in]
$k^T j$ & The same as $\#(k \otimes j)$, the number of ones in $k \otimes j$
   \\[0.07in]
$\Lambda$ & The set of nonnegative real-valued vectors indexed over 
    $\Omega$ whose sum is $1$  \\
& = the set of normalized populations \\
& = the set of probability distributions over $\Omega$\\[0.07in]
$\Omega_u$ & $= \{ k \in \Omega \; : \; u \otimes k = k \}$
   \\[0.07in]
$\Omega_{\overline{u}} \oplus v$ &
$ = \{ j \oplus v \; : \; j \in \Omega_{\overline{u}} \}
  = $ the schema with fixed positions masked by $u$ and specified by $v$
   \\[0.07in]
$x_v^{(u)}$ & $= \sum_{j \in \Omega_{\overline{u}}} x_{j \oplus v}$ 
   (assuming that $\sum_j x_j = 1$).  \\
   & The schema average or sum for the schema $\Omega_{\overline{u}} \oplus v$
   \\[0.07in]
$x^{(u)}$ & The vector of schema averages for the family of schemat
      $\{\Omega_{\overline{u}} \oplus v \:  : \: v \in \Omega_u \}$
   \\[0.07in]
$W$ & The Walsh transform matrix, indexed over $\Omega \times \Omega$.  
     $W_{i,j} = \frac{1}{\sqrt{n}}(-1)^{i^Tj}$
   \\[0.07in]
$W^{(u)}$ & The Walsh transform matrix, indexed over $\Omega_u \times \Omega_u$.
     $W_{i,j}^{(u)} = 2^{-\#u/2}(-1)^{i^Tj}$
   \\[0.07in]
$\widehat{x}$ & $= Wx$, the Walsh transform of normalized population $x$
   \\[0.07in]
$\widehat{x}^{(u)}$ & $= 2^{\#\overline{u}/2}\widehat{x}$, also the
      Walsh transform $W^{(u)}x^{(u)}$ of $x^{(u)}$ with respect to $\Omega_u$
   \\[0.07in]
$\rchi_m$ & The probability that $m \in \Omega$ is used as a crossover mask
   \\[0.07in]
$\mu_m$ & The probability that $m \in \Omega$ is used as a mutation mask
   \\[0.07in]
$\mu_m^{(u)}$ & $\sum_{j \in \Omega_{\overline{u}}} \mu_{k \oplus j}$
   \\[0.07in]
$p_i$ & The probability that bit $i$ is flipped in the mutation step
   \\[0.07in]
$q_i$ & $= 1 - 2p_i$
   \\[0.07in]
$U$ & The matrix indexed over $\Omega \times \Omega$ and defined by
   $U_{j,k} = \mu_{j \oplus k}$
   \\[0.07in]
$U^{(u)}$  & The matrix indexed over $\Omega_u \times \Omega_u$  and defined by 
   $U^{(u)}_{j,k} = \mu^{(u)}_{j \oplus k}$
   \\[0.07in]
\end{tabular}

\end{document}